\documentclass[UTF8]{article}
\usepackage{spconf,amsmath,graphicx}

\title{AIShell-1: An Open-Source Mandarin Speech Corpus and A Speech Recognition Baseline}
\twoauthors
  {Hui Bu}
	{Beijing Shell Shell Technology Co. Ltd,\\
	Beijing, China}
  {Jiayu Du, Xingyu Na, Bengu Wu, Hao Zheng}
	{Independent Researcher\sthanks{The authors performed the work
	while off duty.},\\
	Beijing, China}
\begin{document}
%
\maketitle
\begin{abstract}
An open-source Mandarin speech corpus called AISHELL-1 is released. It is
by far the largest corpus which is suitable for conducting the speech recognition
research and building speech recognition systems for Mandarin.
The recording procedure, including audio capturing devices and environments
are presented in details. The preparation of the related resources, including
transcriptions and lexicon are described. The corpus is released with a Kaldi recipe.
Experimental results implies that the quality of audio recordings and transcriptions
are promising.
\end{abstract}
\section{Introduction}
\label{sec:intro}

Automatic Speech Recognition(ASR) has been an active research topic for several decades.
Most state-of-the-art ASR systems benefit from powerful statistical models, such as
Gaussian Mixture Models(GMM), Hidden Markov Models(HMM) and Deep Neural Networks(DNN)~\cite{dnn}.
These statistical frameworks often require a large amount of high quality data.
Luckily, along with the wide adoption of smart phones, and the emerging market of various
smart devices, real user data are generated world-wide and everyday, hence collecting
data becomes easier than ever before.  Combined with sufficient amount of real data
and supervised-training, statistical approach achieves great success all over the speech industry~\cite{deep}.

However, for legal and commercial reasons, most companies are not willing to share
their data with the public: large industrial datasets are often inaccessible for academic
community, which leads to a divergence between research and industry.  On one hand,
researchers are interested in fundamental problems such as designing new model structures
or beating over-fitting under limited data.  Such innovations and tricks in academic papers
sometimes are proven to be not effective when the dataset gets much larger, different
scales of data lead to different stories.  On the other hand, industrial developers are more
concerned about building products and infrastructures that can quickly accumulate real user
data, then feedback collected data into simple algorithms such as logistic regression and deep learning.  

In ASR community, open-slr project is established to alleviate this problem\footnote{http://www.openslr.org}.
For English ASR, industrial-sized datasets such as Ted-Lium~\cite{ted} and LibriSpeech~\cite{librispeech}
offer open platforms, for both researchers and industrial developers, to experiment and
to compare system performances.  Unfortunately, for Chinese ASR, the only open-source
corpus is THCHS30, released by Tsinghua University, containing 50 speakers, and around
30 hours mandarin speech data~\cite{thchs30}. Generally speaking, Mandarin ASR systems
based on small dataset like THCHS30 are not expected to perform well.  

In this paper, we present AISHELL-1 corpus.  To authors' limited knowledge, AISHELL-1 is
by far the largest open-source Mandarin ASR corpus.  It is released by Beijing Shell-Shell
Company\footnote{http://www.aishelltech.com}, containing 400 speakers and over
170 hours of Mandarin speech data.  More importantly, it is publicly available and is under
Apache 2.0 license.  This paper is organized as below. Section \ref{sec:profile} presents
the recording procedure, including audio capturing devices and environments.  Section \ref{sec:trans}
describes the preparation of the related resources, including transcriptions and lexicon. 
Section \ref{sec:data} explains the final structure of released corpus resources.
In Section \ref{sec:base}, a "drop-in and run" Kaldi recipe is provided as a Mandarin ASR
system baseline.

\section{CORPUS PROFILE}
\label{sec:profile}

AISHELL-1 is a subset of the AISHELL-ASR0009 corpus, which is a 500 hours
multi-channel mandarin speech corpus 
designed for various speech/speaker processing tasks. Speech utterances are recorded via there categories of devices in parallel: 
\begin{enumerate}
\item Microphone: a high fidelity AT2035 microphone with a Roland-R44 recorder working at 44.1 kHz, 16-bit.
\item Android phone: including Samsung NOTE 4, Samsung S6, OPPO A33, OPPO A95s and Honor 6X, working at 16 kHz, 16-bit. 
\item iPhone: including iPhone 5c, 6c and 6s, working at 16 kHz, 16-bit.
\end{enumerate}

\begin{figure}[htp]
\begin{center}
\includegraphics [width=0.5\textwidth] {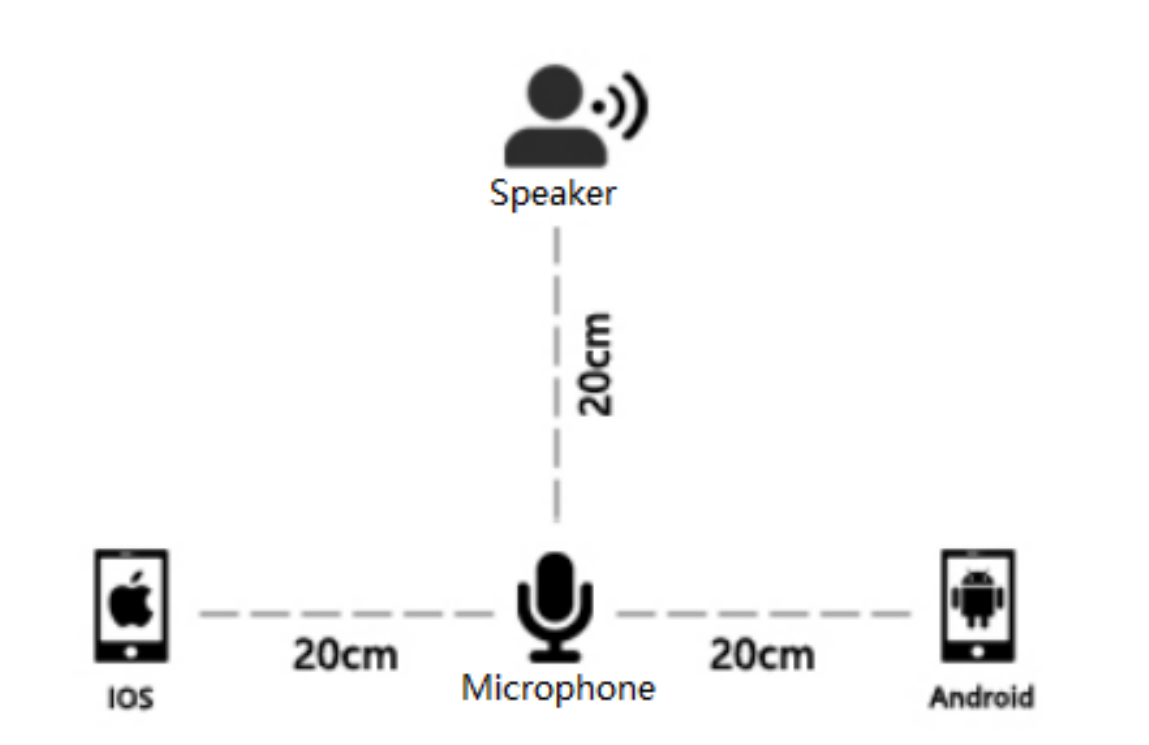}
\caption{Recording setup}
\label{fig:setup}
\end{center}
\end{figure}

The relative position of speaker and devices are shown as Figure \ref{fig:setup}.
The AISHELL-1 database choose high fidelity microphone audio data and re-sampled
to 16 kHz, 16-bit WAV format, which is the mainstream setup for commercial products.

There are 400 participants in the recording, and speakers’ gender, accent, age
and birth-place are recorded as metadata. The gender is 
balanced with 47\% male and 53\% female. As shown in Table \ref{tab:spk}, about 80 percent, of the
speakers are of age 16 to 25. Most speakers come from Northern area of China, detailed distribution is shown in Table \ref{tab:accent}.
The entire corpus includes training, development and
test sets, without speaker overlaping. The details are presented in Section \ref{sec:data}.

\begin{table}[htp]
\caption{Speaker age and gender information}
\begin{center}
\begin{tabular}{|c|c|c|c|}
\hline
Age Range & \#Speakers & Male & Female \\
\hline
16 - 25 yrs & 316 & 140 & 176 \\
26 - 40 yrs & 71 & 36 & 35 \\
$>$ 40 yrs & 13 & 10 & 3 \\
\hline
\end{tabular}
\end{center}
\label{tab:spk}
\end{table}

\begin{table}[htp]
\caption{Speaker accent information}
\begin{center}
\begin{tabular}{|c|c|}
\hline
Accent Area & \#Speakers \\
\hline
North & 333 \\
South & 38 \\
Guangdong-Guangxi-Fujian & 18 \\
Other & 11 \\
\hline
\end{tabular}
\end{center}
\label{tab:accent}
\end{table}

\section{TRANSCRIPTION AND LEXICON}
\label{sec:trans}

The AISHELL-ASR0009 corpus covers common applications such as smart home,
autonomous driving, and the raw text transcriptions are chosen from in 11 domains,
as shown in Table \ref{tab:top}, including 500k commonly used
sentences. The released AISHELL-1 corpus covers 5 of them: ``Finance'',
``Science and Technology'', ``Sports'', ``Entertainments'' and ``News''.
Raw texts are manually filtered to eliminate improper contents involving sensitive
political issues, user privacy, pornography, violence, etc..
Symbols such as $<$, $>$, [, ], \textasciitilde, $/$, $\backslash$, =, etc., are removed.
Long sentences over 25 words are deleted. All text files are encoded in UTF8.

\begin{table}[htp]
\caption{Topics of text}
\begin{center}
\begin{tabular}{|c|c|}
\hline
Topic & \#Sentences \\
\hline
Smart Home Voice Control & 5 \\
POI (Geographic Information) & 30 \\
Music (Voice Control) & 46 \\
Digital Sequence (Voice Control) & 29 \\
TV Play and Film Names & 10 \\
Finance & 132 \\
Science and Technology & 85 \\
Sports & 66 \\
Entertainments & 27 \\
News & 66 \\
English Spelling & 4 \\
\hline
\end{tabular}
\end{center}
\label{tab:top}
\end{table}

In quality checking stage:
\begin{enumerate}
\item data annotators are asked to transcribe speech data, utterances with inconsistent raw text and transcription are removed.
\item Text normalization(TN) is carefully applied towards english words, numbers, name, place, organization, street, shop, brand, examples are:
\begin{itemize}
\item 123 are normalized to yi1 er4 san1 .
\item All the letters or words contained in the URL are capitalized. For example,
the pronunciation content for the ``www.abc.com'', are normalized to
``san1 W dian3 A B C dian3 com''.
\item English abbreviations such as CEO, CCTV are presented in uppercase.
\end{itemize}
\item utterances containing obvious mis-pronunciations are removed.
\end{enumerate}

Besides, A Chinese lexicon is provided in AISHELL-1 corpus. The lexicon is derived from open source lexicon\footnote{https://www.mdbg.net/chinese/dictionary?page=cc-cedict} and covers most of the commonly used Chinese words and characters.
Pronunciations are presented in initial-final syllable.

\section{DATA STRUCTURE}
\label{sec:data}

The corpus includes training set, development set and test sets. Training set contains
120,098 utterances from 340 speakers;
development set contains 14,326 utterance from the 40 speakers; Test set contains
7,176 utterances from 20 speakers. For each speaker, around 360 utterances(about 26
minutes of speech) are released.
Table \ref{tab:struct} provides a summary of all subsets in the corpus.

\begin{table}[htp]
\caption{Data structure}
\begin{center}
\begin{tabular}{|c|c|c|c|}
\hline
Subset & Duration(hrs) & \#Male & \#Female \\
\hline
Training & 150 & 161 & 179 \\
Development & 10 & 12 & 28 \\
Test & 5 & 13 & 7 \\
\hline
\end{tabular}
\end{center}
\label{tab:struct}
\end{table}

\section{SPEECH RECOGNITION BASELINE}
\label{sec:base}

In this section we present a speech recognition baseline released with the corpus
as a Kaldi recipe\footnote{https://github.com/kaldi-asr/kaldi/tree/master/egs/aishell}.
The purpose of the recipe is to demonstrate that this corpus
is a reliable database to conduct Mandarin speech recognition.

\subsection{Experimental setup}
\label{ssec:exp}

The acoustic model (AM) of the ASR system was built largely following the Kaldi
HKUST recipe\footnote{https://github.com/kaldi-asr/kaldi/tree/master/egs/hkust}.
The training started from building an initial Gaussian mixture model-hidden
Markov model (GMM-HMM) system. The acoustic feature consists of two parts,
i.e. 13-dimensional Mel frequency cepstral coefficients (MFCC) and 3-dimensional
pitch features. The selected pitch features are Probability of Voicing (POV) feature obtained
from Normalized Cross Correlation Function (NCCF), log pitch with POV-weighted
mean subtraction over 1.5 second windows, and delta pitch feature computed on
raw log pitch \cite{pitch}. Mean normalization and double deltas are applied on the above features
before feeding into the training pipeline. The GMM-HMM training pipeline is built using tone-dependent
decision trees, meaning that phones with different tonalities as defined in the lexicon
are not clustered together. Maximum likelihood linear transform (MLLT) and speaker
adaptive training (SAT) are applied in the training considering that there is a fair amount
of training data for each of the speakers. The resulting GMM-HMM model has 3, 027
physical p.d.f.s. 

High resolutional (40-dimensional) MFCC and 3- dimensional pitch features are used in
the training of DNN-based acoustic models. Two techniques are applied in DNN training
to enhance acoustic features. The first one is audio augmentation \cite{augment}.
The speaking speed of the training set is perturbed using factor of 0.9 and 1.1, resulting
in a three times larger training set. Besides, the volume of the training data is perturbed
randomly. This technique helps make the DNN model more robust to the tempo and volume
invariances of the testing data. The second technique is i-Vector based DNN adaptation,
which is used to replace mean normalization and double deltas \cite{ivec}. A quarter of the training
data is used to compute a PCA transform and to train a universal background model.
Then all the training data is used to train the i-Vector extractor. Only the MFCCs are used
in the i-Vector extractor training. The estimated i-Vector features are of 100-dimensional.

The DNN model we used was the time delay neural network (TDNN) \cite{tdnn}.
It contained 6 hidden layers, and the activation function was ReLU \cite{relu}. 
The natural stochastic gradient descent (NSGD) algorithm was employed to train
the TDNN \cite{nsgd}. The input feature involved high resolutional MFCC, pitch features,
and the i-Vector feature. A symmetric 4-frame window is applied on MFCC and
pitch features to splice neighboring frames. The output layer consisted of 3, 027 units,
equal to the total number of p.d.f.s in the GMM-HMM model that was trained to bootstrap
the TDNN model.

Lattice-free MMI training is employed for comparison with conventional GMM-HMM bootstrapped
system \cite{lfmmi}. The left-biphone configuration is used and the resulting number of
targets for DNN is 4, 476. The DNN model used in LFMMI training is also TDNN with 6 hidden layers,
and configured to be of the similar number of parameters as the DNN-HMM model.

\subsection{Language model}
\label{ssec:lm}

A trigram language model is trained on 1.3 million words of the training transcripts.
Out-of-vocabulary (OOV) words are mapped into \text{$<$SPOKEN\_NOISE$>$}.
The language model is trained using interpolated Kneser-Ney smoothing and the final
model has 137, 076 unigrams, 438, 252 bigrams and 100, 860 trigrams.

\subsection{Results}
\label{ssec:res}

The results are presented in term of character error rate (CER).  
The base results of GMM-HMM, TDNN-HMM and LFMMI models are shown in
Table \ref{tab:base}. The performances on developing set are better than the testing
set. The performance of LFMMI model is significantly better than TDNN-HMM, indicating
that the corpus has a high transcription quality. Audio quality can be reflected by the
performance on totally difference data from the training set. Thus we evaluate the models
on the mobile recording channel and THCHS30 testing set.

\begin{table}[htp]
\caption{Baseline results}
\begin{center}
\begin{tabular}{|c|c|c|}
\hline
Model & CER of dev & CER of test \\
\hline
MLLT+SAT & 10.43\% & 12.23\% \\
TDNN-HMM & 7.23\% & 8.42\% \\
LFMMI & 6.44\% & 7.62\% \\
\hline
\end{tabular}
\end{center}
\label{tab:base}
\end{table}

\subsubsection{Decoding the mobile recordings}
\label{ssec:mobres}

The parallel testing recordings using Android and iOS devices are selected from
the AISHELL-ASR0009 corpus, and they are used
to evaluate the performance of the AISHELL-1 model on less fidelity devices.
Results are shown in Table~\ref{tab:mobile}. Device mismatch results in
significant performance loss. However, stronger acoustic models improves the
performance on such less fidelity devices.

\begin{table}[htp]
\caption{Mobile recording results}
\begin{center}
\begin{tabular}{|c|c|c|}
\hline
Model & CER of iOS & CER of Android \\
\hline
MLLT+SAT & 12.64\% & 11.88\% \\
TDNN-HMM & 12.42\% & 10.81\% \\
LFMMI & 10.90\% & 10.09\% \\
\hline
\end{tabular}
\end{center}
\label{tab:mobile}
\end{table}

\subsubsection{Decoding the THCHS30 test set}
\label{ssec:thchsres}

The performance of AISHELL-1 model on testing cases of an unrelated
language model domain than the training set reflects the overall quality
of the corpus. Table~\ref{tab:thchs} shows that stronger acoustic models
performs better on an unrelated domain, indicating that the corpus is phonetically
covered and an adapted language model will fill the performance gap.

\begin{table}[htp]
\caption{THCHS30 testing set results}
\begin{center}
\begin{tabular}{|c|c|}
\hline
Model & CER \\
\hline
MLLT+SAT & 32.23\% \\
TDNN-HMM & 28.15\% \\
LFMMI & 25.00\% \\
\hline
\end{tabular}
\end{center}
\label{tab:thchs}
\end{table}

\section{CONCLUSIONS}
\label{sec:con}

An open-source Mandarin corpus is released\footnote{http://www.openslr.org/33/}.
To our best knowledge, it is the largest academically free data set for Mandarin
speech recognition tasks. Experimental results are presented using the Kaldi recipe
published along with the corpus. The audio and transcription qualities are promising
for constructing speech recognition systems for Mandarin.

\vfill
\pagebreak

\bibliographystyle{IEEEbib}
\bibliography{strings,refs}

\end{document}